\documentclass[]{fairmeta}
\usepackage{xcolor}
\usepackage{colortbl}
\usepackage{amsfonts}
\usepackage{xspace}
\usepackage{pifont}
\usepackage{overpic}
\newcommand{\xmark}{\ding{55}}  % Add this in the preamble of your document
\newcommand\our{{\fontfamily{lmtt}\selectfont MetaQuery}\xspace}
\newcommand\ours{{\fontfamily{lmtt}\selectfont MetaQueries}\xspace}

\usepackage{arydshln}
\usepackage{wrapfig}
\usepackage{subcaption}

\newcommand{\canvas}[1]{
    \vspace{-0.1cm}
    \begin{tcolorbox}[
 colback=white!90!gray,     % Soft ivory background
 colframe=teal!60!black,     % Deep navy frame color
 arc=5pt,                    % Less rounded corners for a modern look
 boxsep=5pt,                 % Inner margin between text and frame
 left=10pt,                  % Left margin within the box
 right=10pt,                 % Right margin within the box
 top=2pt,                    % Top margin within the box
 bottom=2pt,                 % Bottom margin within the box
 boxrule=0.8pt,              % Frame thickness
 drop shadow=gray!50!white,  % Subtle shadow effect for depth
 enhanced jigsaw             % Allows for the shadow effect and better arcs
 ]
    \vspace{-0.1cm}
 #1
    \vspace{-0.1cm}
    \end{tcolorbox}
    \vspace{-0.3cm}
}

\title{Transfer between Modalities with MetaQueries}

\author[1,2]{Xichen Pan}
\author[1,\dagger]{Satya Narayan Shukla}
\author[1]{Aashu Singh}
\author[1]{Zhuokai Zhao}
\author[1]{Shlok Kumar Mishra}
\author[1]{Jialiang Wang}
\author[1]{Zhiyang Xu}
\author[1]{Jiuhai Chen}
\author[1]{Kunpeng Li}
\author[1]{Felix Juefei-Xu}
\author[1,\dagger]{Ji Hou}
\author[2,\dagger]{Saining Xie}

\affiliation[1]{Meta}
\affiliation[2]{New York University}

\contribution[\dagger]{Equal advising}

\abstract{
Unified multimodal models aim to integrate understanding (text output) and generation (pixel output), but aligning these different modalities within a single architecture often demands complex training recipes and careful data balancing. We introduce \ours, a set of learnable queries that act as an efficient interface between autoregressive multimodal LLMs (MLLMs) and diffusion models. \ours connects the MLLM's latents to the diffusion decoder, enabling knowledge-augmented image generation by leveraging the MLLM's deep understanding and reasoning capabilities. Our method simplifies training, requiring only paired image-caption data and standard diffusion objectives. Notably, this transfer is effective even when the MLLM backbone remains frozen, thereby preserving its state-of-the-art multimodal understanding capabilities while achieving strong generative performance. Additionally, our method is flexible and can be easily instruction-tuned for advanced applications such as image editing and subject-driven generation.
}

\date{\today}
\correspondence{\email{satyanshukla@meta.com}, \email{jihou@meta.com}, \email{saining.xie@nyu.edu}}

\metadata[Project Page]{\url{https://xichenpan.com/metaquery}}

\begin{document}

\maketitle

\section{Introduction}
The quest for unified multimodal models capable of both deep understanding (typically resulting in textual outputs) and rich generation (resulting in pixel outputs) holds immense promise. Such systems could unlock synergistic capabilities~\citep{gpt4oimagegeneration, geminiimagegeneration}, where understanding informs generation and vice versa. However, effectively connecting these different output modalities poses considerable challenges---\emph{e.g.} how do we effectively transfer the latent world knowledge from the autoregressive multimodal LLM to the image generator? Although significant progress has been made, most published approaches~\citep{seedx,emu,metamorph,lavit,lwm,chameleon,showo,emu3,janus,januspro,dreamllm,transfusion,lmfusion} rely on carefully tuning base multimodal LLMs (MLLMs) to handle both understanding and generation tasks. This involves complex architectural design, data/loss balancing, multiple training stages, and other complex training recipes---without these, optimizing one capability could compromise the other. 

In this paper, we aim to deliver the promise of unified models via a simpler philosophy: \emph{Render unto diffusion what is generative, and unto LLMs what is understanding.} In other words, instead of building a monolithic system from scratch, we focus on effectively transferring capabilities between state-of-the-art, pre-trained models specialized for different output modalities. To operationalize this, we keep MLLMs frozen so they can focus on what they do best---understanding---while entrusting image generation to diffusion models. We then demonstrate that even under this frozen condition, the MLLM's inherent world knowledge, strong reasoning, and in-context learning capabilities can indeed be transferred to image generation, provided the right architectural bridge is in place.

\begin{figure}
    \centering
    \includegraphics[width=0.9\linewidth]{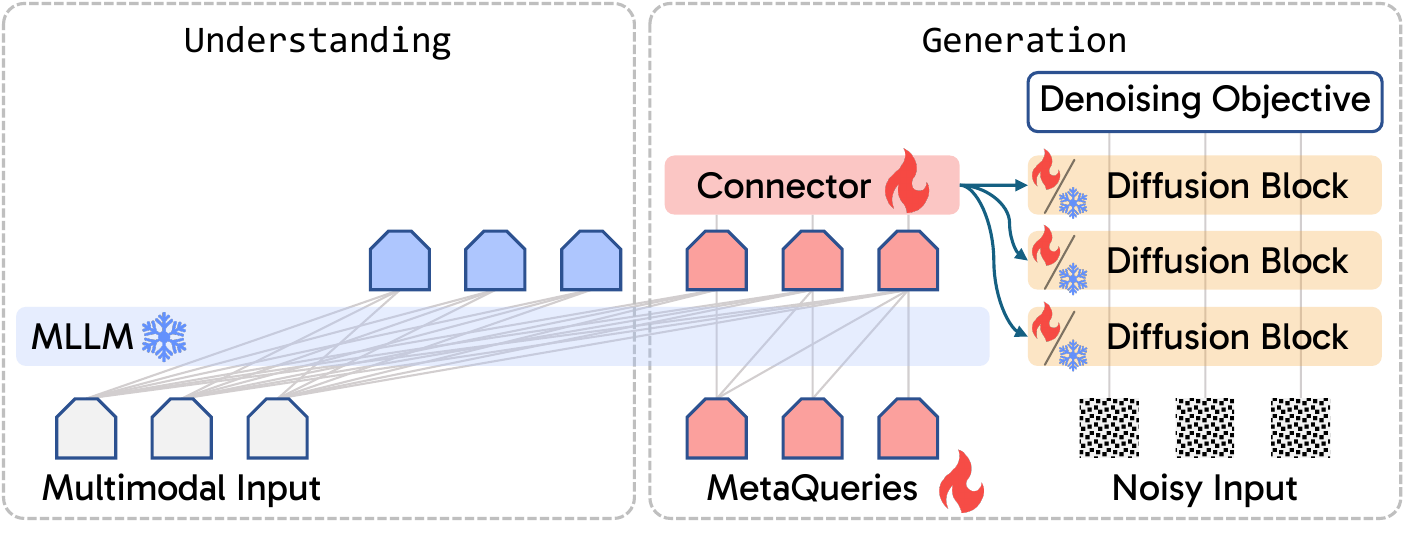}
    \caption{Overview of our model. \colorbox[HTML]{B3C5F9}{Blue tokens} maintain SOTA multimodal understanding; \colorbox[HTML]{EEA49F}{\ours} are learnable queries that directly applied to frozen MLLMs to query out conditions for generation. The model is tuned using only denoising objective with paired data. The generative diffusion models can be either frozen or further instruction-tuned for advanced generation tasks.}
    \label{fig:metaquery}
\end{figure}

However, leveraging an MLLM---especially a frozen one---for both multimodal understanding and generation is far from straightforward. Although (frozen) LLMs have shown good performance as conditional text encoders in text-to-image generation \citep{luminanext, sana, lidit}, they are not compatible with many desired tasks in unified modeling, such as in-context learning or producing multimodal, interleaved output. The {architectural bridge} we design in this work is \our (Figure~\ref{fig:metaquery}). \our feeds a set of learnable queries directly into a frozen MLLM to extract multimodal conditions for multimodal generation. Our experiments reveal that, even without fine-tuning or enabling bi-directional attention, the frozen LLM serves as a powerful feature resampler~\citep{flamingo}, producing high-quality conditions for multimodal generation. Training unified models with \ours requires only a modest amount of paired image-caption data to connect these prompted conditions to any conditional diffusion model. Because the entire MLLM stays intact for understanding, the training objective remains the original denoising objective---just as efficient and stable as fine-tuning a diffusion model.

More specifically, previous unified models aim to train a single autoregressive transformer backbone to jointly model $p(\texttt{text}, \texttt{pixels})$. In contrast, we choose to use a $\texttt{token} \rightarrow [\texttt{transformer}] \rightarrow [\texttt{diffusion}] \rightarrow \texttt{pixels}$ paradigm, which might share a high-level philosophy with the concurrent GPT-4o image generation system, as hinted at by~\cite{gpt4oimagegeneration}. This approach composes the MLLM's autoregressive prior with a powerful diffusion decoder, directly leveraging the frozen MLLM's strong capability in modeling compressed semantic representations, thus avoiding the more challenging task of directly generating pixels.

To validate our approach, we conduct a series of controlled experiments, showing that \our\footnote{For simplicity, we also use \our to represent our method.} outperforms the use of a frozen MLLM purely as a conditional text encoder for image generation. Moreover, \our can match the performance of fully tuning the MLLM backbone, yet it is significantly more efficient. We also systematically investigate the training strategy, including the number of tokens and architectural configurations. With just 25M publicly available image-caption pairs, we are able to train a family of unified models that not only preserves state-of-the-art (SOTA) performance in image understanding, but also achieves SOTA-level results in text-to-image generation across multiple benchmarks.

The promise of unified modeling goes beyond handling multimodal understanding and text-to-image generation in parallel. A deeper synergy is expected---one that taps into advanced MLLM abilities like reasoning, internal knowledge, multimodal perception, and in-context learning to enhance generation. Our results show that our method draws on the frozen MLLM's commonsense knowledge, achieving SOTA visual-commonsense generation on the CommonsenseT2I benchmark~\citep{commonsenset2i}. Our approach also harnesses the built-in reasoning and in-context learning capabilities of frozen MLLMs, producing images from complex prompts---such as generating the United States flag in response to ``\emph{The national flag of the country where Yellowstone National Park is located.}'' (See Figure~\ref{fig:commonsense} for examples.) We also benchmark this type of world knowledge reasoning capability on WISE~\citep{wise} and demonstrate SOTA performance.

Finally, by connecting, preserving, and enhancing multimodal input with \ours and a frozen MLLM backbone, our model can be further instruction-tuned for advanced generation tasks such as image editing and subject-driven generation. We show that this can be achieved both efficiently and effectively using a scalable data curation pipeline that directly leverages naturally occurring image pairs from web corpora, instead of depending on human-created pairs or synthetically generated data~\citep{Instructpix2pix, instructimagen, Omnigen}. This natural supervision surprisingly unlocks several new capabilities beyond subject-driven generation, such as visual association and logo design (see Figure~\ref{fig:subjectdriven} for examples).

In summary, we explore a simple yet underexplored alternative to unified multimodal modeling. Our method, \our, bridges frozen MLLM backbones and diffusion models. Experiments show that this framework delivers all the capabilities once thought to require MLLM fine-tuning while being much easier to train. The main results and findings in this paper include:
\begin{itemize}
    \item With \our and frozen MLLM backbones, we maintain SOTA multimodal understanding performance while enabling SOTA-level multimodal generation.
    \item \our can transfer the capabilities of MLLMs for reasoning- and knowledge-augmented image generation.
    \item \our can extract highly detailed visual conditions beyond semantic similarity from frozen MLLMs, enabling image reconstruction and editing tasks.
    \item Our method can be easily instruction-tuned even with a frozen MLLM backbone, enabling advanced multimodal generation tasks like subject-driven generation.
\end{itemize}

\section{Related Work}

\paragraph{Unified understanding and generation models.}
Next-token prediction has proven to be an effective approach for training models to understand language~\citep{bert, gpt} and multimodal content~\citep{llava}. Recently, the community has witnessed numerous efforts to extend the success of multimodal understanding~\citep{llava} to multimodal generation by training LLM backbones to generate images at the same time. However, unlike adapting text-only LLMs~\citep{llama} to understand multimodal content with one single next text token prediction objective~\citep{llava}, generating multimodal content requires a different set of training objectives. SEED-X~\citep{seedx}, Emu~\citep{emu}, and MetaMorph~\citep{metamorph} learn to regress image features; LaVIT~\citep{lavit}, LWM~\citep{lwm}, Chameleon~\citep{chameleon}, Show-o~\citep{showo}, EMU3~\citep{emu3}, and Janus~\citep{janus,januspro} auto-regressively predict next visual tokens; and DreamLLM~\citep{dreamllm}, Transfusion~\citep{transfusion} employ diffusion objectives. However, these approaches necessitate tuning LLMs for generating both modalities, naturally posing challenges in multi-task balancing.

\paragraph{Unified models with frozen LLMs.}
Several studies have explored the use of frozen LLMs for multimodal understanding and generation. For instance, LMFusion~\citep{lmfusion} trains image generation expert feed-forward networks (FFNs) and query-key-value (QKV) modules in parallel with a frozen LLM backbone to deeply fuse input conditions and denoise visual outputs. However, this approach offers limited flexibility as it shares the same architecture as specific LLM backbones and requires training a separate set of generative modules for every single LLM backbone. This not only imposes more computational burden but also restricts the ability to leverage powerful pre-trained generative models. An earlier work, GILL~\citep{gill}, investigates feeding learnable tokens into frozen MLLMs. It employs a combined contrastive loss and regression loss for image retrieval and generation, rather than directly employing the denoising objective for more efficient training. Its application is restricted to contextual image generation and it does not systematically explore the impact of frozen MLLMs and learnable queries.

\section{\our}
In this work, we propose \our, which losslessly augments understanding-only MLLMs with multimodal generation capabilities while preserving their original architecture designs and parameters intact. We carefully analyze the impact of applying \our on image generation performance. Results show that a frozen MLLM can provide strong conditions for multimodal generation.

\subsection{Architecture}
\our bridges frozen MLLMs with diffusion models. We use randomly initialized learnable queries $\mathcal{Q}\in \mathbb{R}^{N\times D}$ to query out the conditions $\mathcal{C}$ for generation. $N$ is the number of queries and $D$ is the dimension of the queries, which is the same as the MLLM hidden dimension. 
For simplicity and compatibility, we continue to use causal masking for the entire sequence rather than specifically enabling full attention for $\mathcal{Q}$.
The conditions $\mathcal{C}$ are then fed into a trainable connector to align with the input space of text-to-image diffusion models. These models can be arbitrary as long as they have a conditional input interface; we simply replace its original condition with our $\mathcal{C}$. The whole model is trained with the original generation objective on paired data. In this paper, we focus on image generation tasks, but the model can be easily extended to other modalities like audio, video, 3D, and more.

\begin{table}[!t]
    \small
    \centering
    \begin{tabular}{lcccc}
        
        \textbf{Methods} & \textbf{\# of Tokens} & \textbf{MJHQ-30K FID $\downarrow$} & \textbf{GenEval $\uparrow$} & \textbf{DPG-Bench $\uparrow$} \\
        \midrule
 LLM last layer embedding$^*$ & - & 7.49 & 0.55 & 78.41 \\
 Random queries & 64 & 8.59 & 0.35 & 54.81 \\
 Learnable queries & 64 & 7.43 & \cellcolor{green!10}0.56 & 75.35 \\
 Learnable queries & 512 & \cellcolor{green!10}7.34 & \cellcolor{green!10}0.56 & \cellcolor{green!10}78.43 \\

    \end{tabular}
    \caption{Study on different conditions for image generation. $^*$ denotes the embeddings of input tokens. Learnable queries achieve comparable performance to using all hidden states and can even surpass them with more tokens.}
    \label{tab:learnable_queries}
\end{table}

\begin{table}[!t]
    \small
    \centering
    \begin{tabular}{lccccc}
        
        \textbf{Methods} & \textbf{Train LLM} & \textbf{Train DiT} & \textbf{MJHQ-30K FID $\downarrow$} & \textbf{GenEval $\uparrow$} & \textbf{DPG-Bench $\uparrow$} \\
        \midrule
 MLLM tuning & \checkmark & \xmark & 7.75 & 0.58 & 78.97 \\
 E2E tuning & \checkmark & \checkmark & 6.28 & \cellcolor{green!10}0.61 & \cellcolor{green!10}79.39 \\
 Frozen MLLM & \xmark & \xmark & 7.43 & 0.56 & 75.35 \\
 Frozen MLLM & \xmark & \checkmark & \cellcolor{green!10}6.06 & \cellcolor{green!10}0.61 & 76.66 \\
    \end{tabular}
    \caption{Study on strategies for adapting MLLMs. The methods without training LLM do not suffer from multimodal understanding degradation. Frozen MLLM achieves comparable performance to full MLLM tuning, with slightly lower prompt alignment but slightly improved visual quality.}
    \label{tab:frozen_mllm}
\end{table}

\subsection{Design Choices}
\label{sec:design_choices}
The proposed architecture involves two design choices: using \textbf{learnable queries} and keeping the \textbf{MLLM backbone frozen}. We explain the reasons why we adopted these choices and how they impact performance. For all experiments, unless otherwise specified, we use the same frozen LLaVA-OneVision-0.5B~\citep{llavaov} MLLM backbone, frozen Sana-0.6B~\citep{sana} diffusion model in 512 resolution, learnable queries with $N=64$ tokens, and a connector with a 24-layer transformer encoder. All models are trained on 25M publicly available image caption pairs for 4 epochs. We report FID score~\citep{fid} on MJHQ-30K~\citep{playgroundv2p5} for visual aesthetic quality, and GenEval~\citep{geneval} and DPG-Bench~\citep{dpg} (both without prompt rewriting) for prompt alignment, respectively.

\paragraph{Learnable queries.}
Many models like Lumina-Next~\citep{luminanext}, Sana~\citep{sana}, and Kosmos-G~\citep{kosmosg} use the (M)LLM's last layer embedding of input tokens as image generation conditions. However, this approach is not ideal for unified models as it is not compatible with many desired tasks in unified modeling, such as in-context learning or producing multimodal, interleaved output (we provide more discussion and comparison with \our in Section~\ref{sec:discussion}). As shown in Table~\ref{tab:learnable_queries}, using learnable queries with just $N=64$ tokens achieves image generation quality comparable to that of utilizing the last layer embedding of input tokens. While random queries produce acceptable FID scores, they struggle with prompt alignment, highlighting the importance of learnable queries. Additionally, since the last layer embedding setting naturally comes with a longer sequence length, we also tested learnable queries with $N=512$ tokens, which further improves performance and even outperforms the last layer embedding approach.

\paragraph{Frozen MLLM.}
Existing unified models train MLLMs to jointly model $p(\texttt{text}, \texttt{pixels})$, resulting in a more complicated training process and even downgraded understanding performance. \our keeps the original MLLM architecture and parameters intact to preserve SOTA understanding capabilities. However, for multimodal generation, a key concern is whether \our's performance with significantly fewer tunable parameters would be substantially worse than methods with full MLLM tuning. As shown in Table~\ref{tab:frozen_mllm}, frozen MLLMs achieve comparable performance to full MLLM tuning, with slightly lower prompt alignment but slightly improved visual quality. Tuning DiT can further improve performance for both settings. This suggests that \our is another possible training strategy, one that is simpler but also effective, as an alternative to fine-tuning the entire MLLM.

\subsection{Training Recipe}
Based on insights from our design choices, we further study key training options for the two main components of \our: learnable queries and connectors. This study examines the number of tokens and connector design. Unless otherwise specified, all experiments in this section use the same setup as described in Section~\ref{sec:design_choices}.

\begin{figure}[!t]
    \centering
    \begin{subfigure}[b]{0.50\linewidth}
        \centering
        \includegraphics[width=\linewidth]{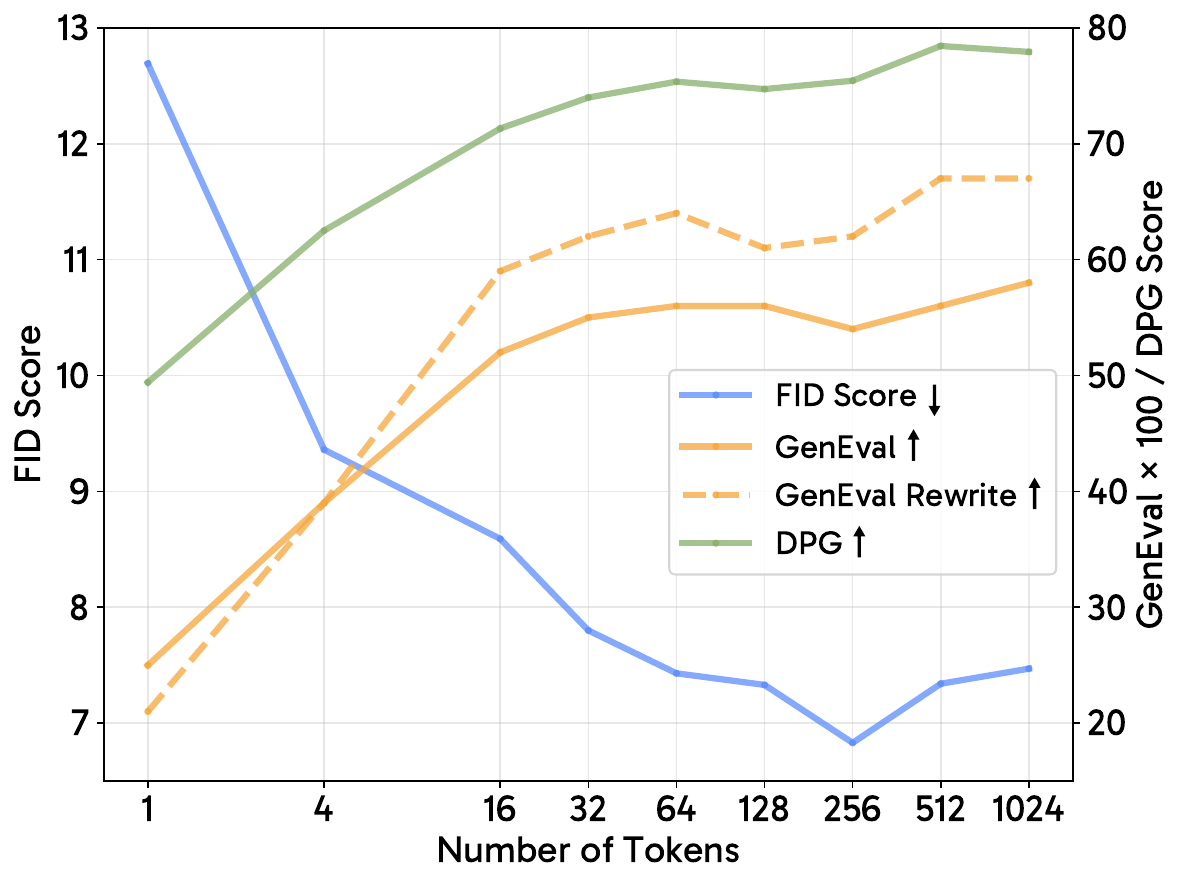}
        \caption{Text-to-image results.}
        \label{fig:num_tokens}
    \end{subfigure}
    \hfill
    \begin{subfigure}[b]{0.49\linewidth}
        \centering
        \includegraphics[width=\linewidth]{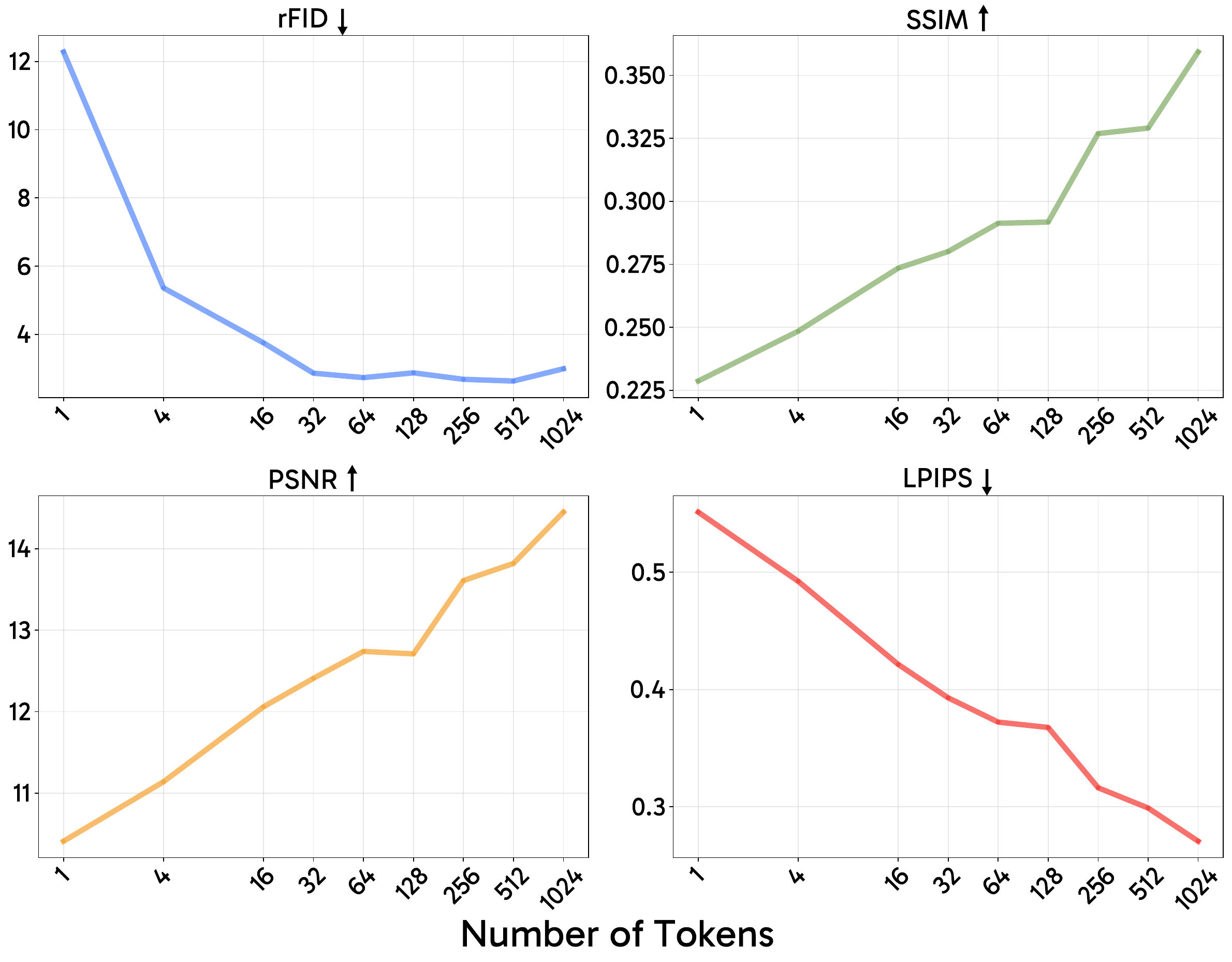}
        \caption{Image reconstruction results.}
        \label{fig:num_of_tokens_rec}
    \end{subfigure}
    \caption{Study on the scaling of token numbers. As the number of tokens increases, text-to-image prompt alignment and image reconstruction results consistently improve.}
    \label{fig:num_tokens_all}
\end{figure}

\begin{figure}[!t]
    \centering
    \begin{overpic}[width=\linewidth]{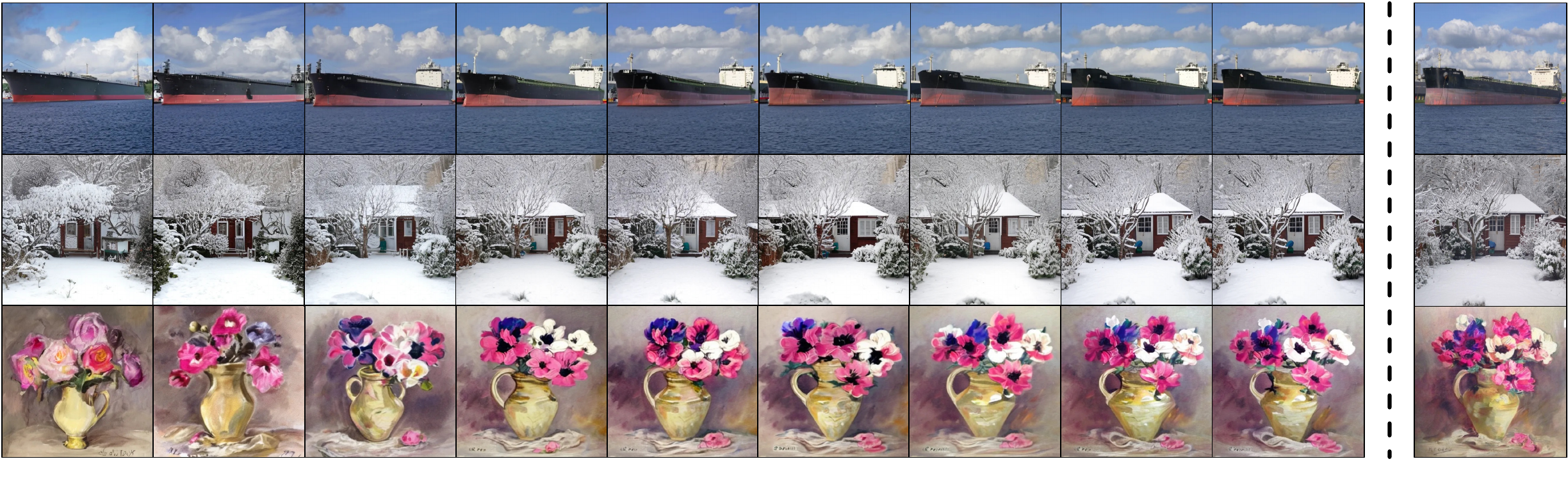}
        \put(2.5, 0) {\scriptsize $N=1$}
        \put(12.5, 0) {\scriptsize $N=4$}
        \put(21.5, 0) {\scriptsize $N=16$}
        \put(31.2, 0) {\scriptsize $N=32$}
        \put(41, 0) {\scriptsize $N=64$}
        \put(50, 0) {\scriptsize $N=128$}
        \put(59.7, 0) {\scriptsize $N=256$}
        \put(69.5, 0) {\scriptsize $N=512$}
        \put(78.7, 0) {\scriptsize $N=1024$}
        \put(91, 0) {\scriptsize Real Image}
    \end{overpic}
    \caption{Visaul samples for image reconstruction with different numbers of tokens.}
    \label{fig:num_of_tokens_rec_samples}
\end{figure}
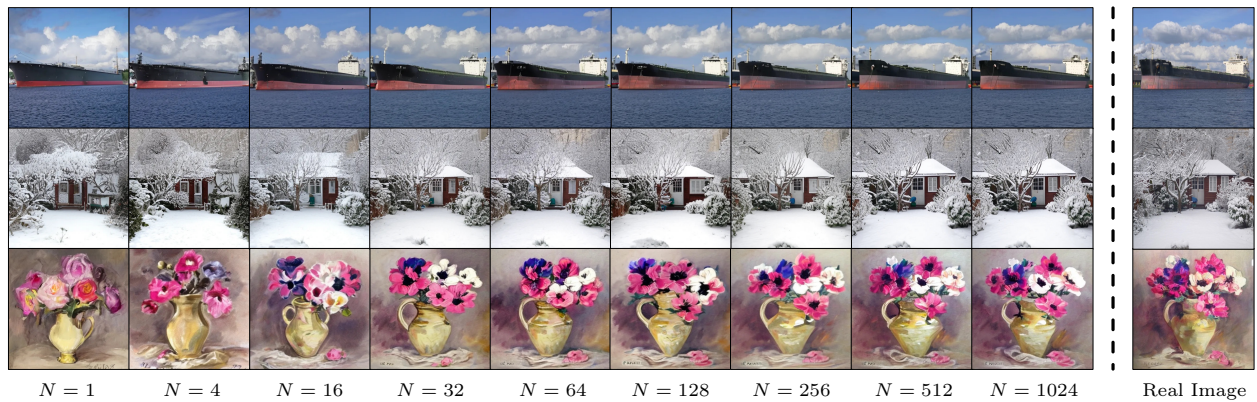

\paragraph{Number of tokens.}
Many works~\citep{nextgpt, kosmosg, seedx} have employed learnable queries for condition extraction. However, they either set the number of tokens to match the fixed input sequence length of the image decoder (e.g., $N=77$ for the CLIP~\citep{clip} text encoder in Stable Diffusion v1.5~\citep{sd1p5}), or use an arbitrary fixed number like $N=64$ without further investigation. Given that modern diffusion models like Lumina-Next~\citep{luminanext} and Sana~\citep{sana} naturally accept variable-length conditions, determining the optimal number of tokens for learnable queries is crucial. In Figure~\ref{fig:num_tokens_all}, we provide a careful study of the number of tokens and observe promising scalability of \ours. For text-to-image generation, visual quality begins to converge after 64 tokens, while more tokens consistently yield better prompt alignment. This is more evident for long captions, as GenEval with rewritten prompts increases more rapidly as the number of tokens increases. For image reconstruction, we observe that more tokens consistently improve the quality of reconstructed images (visual samples can be found in Figure~\ref{fig:num_of_tokens_rec_samples}). In our later experiments, we set the number of tokens to $N=256$ for all models, as it achieves a good balance between performance and efficiency.

\begin{table}[!t]
    \small
    \centering
    \begin{tabular}{lccccccc}
        
        \textbf{Architecture} & \textbf{\# of Layers} & \textbf{Dims} & \textbf{\# of Params} & \textbf{Rel. Wall Time} & \textbf{MJHQ-30K FID $\downarrow$} & \textbf{GenEval $\uparrow$} & \textbf{DPG-Bench $\uparrow$} \\
        \midrule
        % 1368
 Proj-Enc & 6 & 2304 & 517M & 1.06x & 7.80 & 0.53 & 73.37 \\
 Proj-Enc & 24 & 2304 & 2046M & 1.23x & \cellcolor{green!10}7.41 & 0.51 & 73.75 \\
 Enc-Proj & 6 & 896 & 84M & 1x & 7.73 & 0.49 & 71.39 \\
 Enc-Proj & 24 & 896 & 316M & 1.06x & 7.43 & \cellcolor{green!10}0.56 & \cellcolor{green!10}75.35 \\
        
    \end{tabular}
    \caption{Study on connector design. Aligning the conditions first in the same dimension as the MLLM hidden states (Enc-Proj) is more effective and parameter-efficient.}
    \label{tab:connector_design}
\end{table}

\paragraph{Connector design.}
The connector is another important component in \our. We use the same architecture as the Qwen2.5~\citep{qwen2.5} LLM, but enable bi-directional attention for the connector. We study two different designs: Projection Before Encoder (Proj-Enc) and Projection After Encoder (Enc-Proj). Proj-Enc first projects the conditions into the input dimension of the diffusion decoder, then uses a transformer encoder to align the conditions. On the other hand, Enc-Proj first uses a transformer encoder to align the conditions in the same dimension as the MLLM hidden states, then projects the conditions into the input dimension of the diffusion decoder. As shown in Table~\ref{tab:connector_design}, the Enc-Proj design achieves better performance than the Proj-Enc design while having fewer parameters.

\section{Model Training}
\label{sec:model_training}
\begin{wrapfigure}[20]{r}{0.63\linewidth}
    \centering
    \vspace{-1.5em}
    \includegraphics[width=\linewidth]{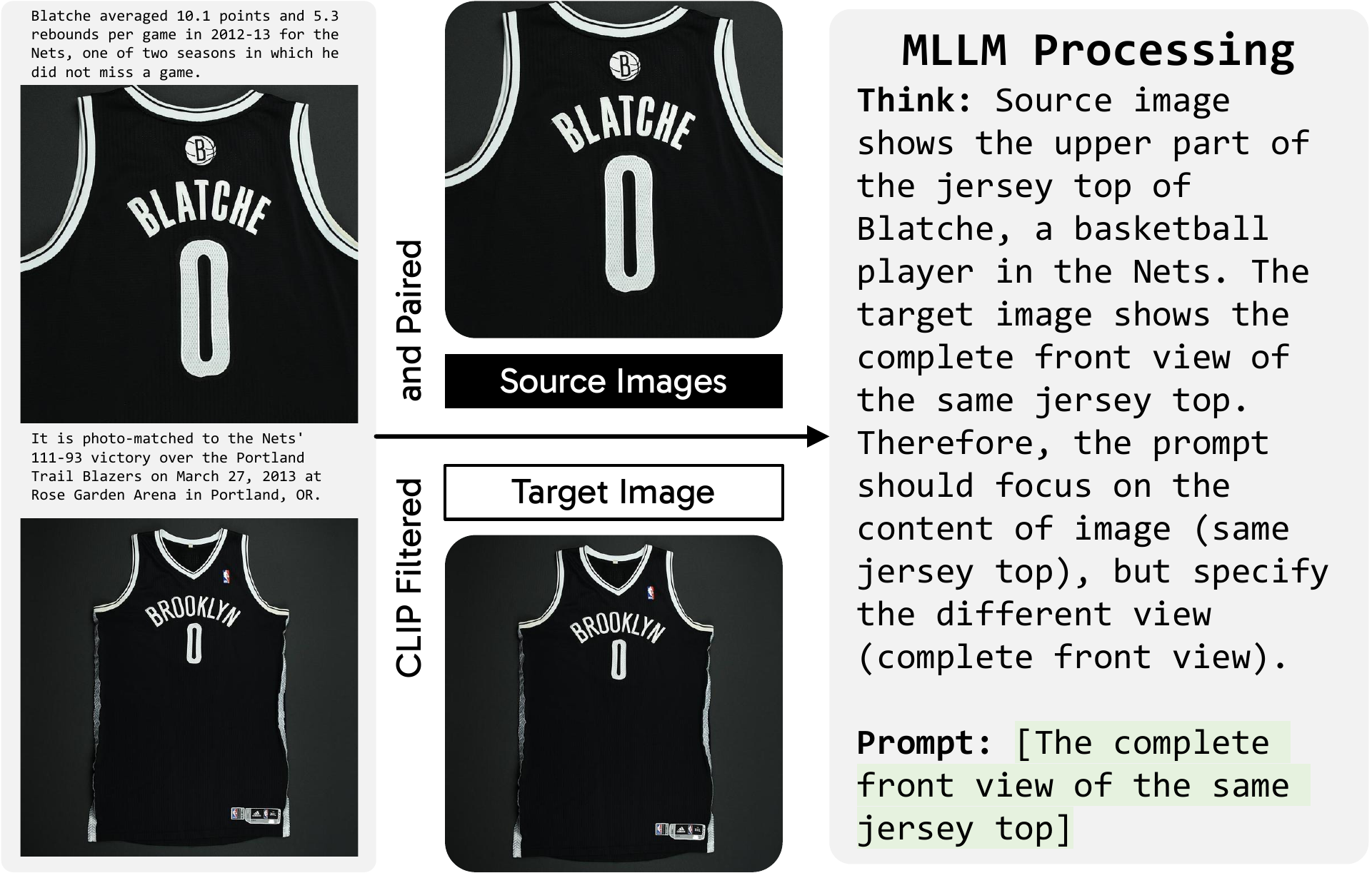}
    \caption{Overview of instruction tuning data curation pipeline. We group images from web corpora based on caption similarity using the SigLIP~\citep{siglip} model, then construct instruction-tuning data from these image pairs using an MLLM.}
    \label{fig:instruction_tuning_data}
\end{wrapfigure}
We train \our in two stages: the pre-training stage and the instruction tuning stage. Both training stages keep MLLMs frozen and fine-tune learnable queries, connectors, and diffusion models. We use three different MLLM backbones for different sizes: Base (LLaVA-OneVision 0.5B~\citep{llavaov}), Large (Qwen2.5-VL 3B~\citep{qwen2p5vl}), and X-Large (Qwen2.5-VL 7B~\citep{qwen2p5vl}). We set the number of tokens to $N=256$ for all models, and utilize a 24-layer connector with Enc-Proj architecture. For image generation heads, we tested two different diffusion models: Stable Diffusion v1.5~\citep{sd1p5} and Sana-1.6B~\citep{sana}.

\paragraph{Pre-training.}
We pre-train our model on 25M publicly available image-caption pairs for 8 epochs with a learning rate of 1e-4 and a global batch size of 4096. The learning rate follows a cosine decay schedule with a 4,000-step warmup period before gradually decreasing to 1e-5.

\paragraph{Instruction tuning.}
Furthermore, in this work, we rethink the data curation process for instruction tuning in image generation. All current methods rely on expert models to generate target images from source images and instructions~\citep{seedx, Omnigen, instructimagen}. However, this approach is limited in scalability and may introduce biases, as the available expert models cover only a narrow range of image transformations. Inspired by MagicLens~\citep{Magiclens}, we construct instruction-tuning data using naturally occurring image pairs in web corpora. These corpora contain rich multimodal contexts with interleaved text and images on related subjects or topics. These image pairs often exhibit meaningful associations and specific relationships spanning a broad spectrum, from direct visual similarities to more subtle semantic connections (as shown in Figure~\ref{fig:instruction_tuning_data}). Such naturally occurring image pairs provide excellent and diverse supervision signals for instruction tuning. Based on this observation, we developed a data construction pipeline that mines image pairs and leverages MLLMs to generate open-ended instructions that capture their inter-image relationships. First, we collect grouped images from mmc4~\citep{mmc4} core fewer-faces subset, where each image is accompanied by a caption. Using SigLIP~\citep{siglip}, we cluster images with similar captions (allowing up to 6 images per group, with a similarity threshold of 0.5). In each group, the image with minimum average similarity to the others is designated as the target, while the remaining images serve as source images. This process yields a total of 2.4M image pairs. Finally, we employ Qwen2.5-VL 3B~\citep{qwen2p5vl} to generate instructions for each pair, describing how to transform the source images into the target image (See Appendix~\ref{app:details_in_data_curation} for the detailed MLLM prompt). We experimented with instruction-tuning our Base size model on the proposed 2.4M dataset for 3 epochs, using the same learning rate schedule as in pre-training and a batch size of 2048.

\section{Experiments}
In this section, we first evaluate \our on various multimodal understanding and text-to-image generation benchmarks (Section~\ref{sec:img_understanding_generation}). We demonstrate that \our can be trained to reconstruct input images (Section~\ref{sec:img_reconstruction}). This image reconstruction capability can be easily transferred to perform image editing (Section~\ref{sec:img_editing}). Furthermore, we show that \our can be instruction-tuned to perform zero-shot subject-driven generation (Section~\ref{sec:subject_driven}). By leveraging our approach for collecting instruction tuning data from naturally existing image pairs, we also reveal that \our can unlock novel capabilities like visual association and logo design (also in Section~\ref{sec:subject_driven}). Additionally, we demonstrate that \our can benefit from the internal knowledge and reasoning capabilities of the frozen MLLM, overcoming common failures exhibited by other generation models (Section~\ref{sec:reasoning_and_knowledge_augmented_generation}). Finally, we discuss the impact of different MLLM backbones and compare \our's behavior with the baseline that uses MLLM last layer embeddings (Section~\ref{sec:discussion}).

\begin{table*}[!t]
    % \fontsize{7pt}{8pt}\selectfont
    \scriptsize
    \centering
    \setlength{\tabcolsep}{4pt}
    \renewcommand{\arraystretch}{1.2}
    \begin{tabular}{llccccc|cccc}
        \textbf{Methods} & \textbf{Base (M)LLM} & \textbf{MME-P} & \textbf{MMB} & \textbf{SEED} & \textbf{MMMU} & \textbf{MM-Vet} & \textbf{COCO FID $\downarrow$} & \textbf{MJHQ FID $\downarrow$} & \textbf{GenEval $\uparrow$} & \textbf{DPG-Bench $\uparrow$} \\
        \midrule        
 Emu & LLaMA 13B & - & - & - & - & - & 11.66 & - & - & - \\
 DreamLLM & Vicuna 7B & - & - & - & - & 36.6 & 8.46 & - & - & - \\
 Chameleon & From Scratch 7B & - & - & - & 22.4 & 8.3 & 26.74 & - & 0.39 & - \\
 Show-o-512 & Phi-1.5 1.3B & 1097.2 & - & - & 26.7 & - & 9.24 & 15.18 & 0.68 & - \\
 VILA-U & LLaMA-2 7B & 1401.8 & - & 59.0 & - & 33.5 & - & 7.69 & - & - \\
 Emu3 & From Scratch 7B & - & 58.5 & 68.2 & 31.6 & 37.2 & 12.80 & - & 0.66$^\dagger$ & 80.60 \\
 MetaMorph & LLaMA-3 8B & - & 75.2 & 71.8 & - & - & 11.8 & - & - & - \\
 TokenFlow-XL & Qwen-2.5 14B & 1551.1 & 76.8 & 72.6 & 43.2 & 48.2 & - & - & 0.63$^\dagger$ & 73.38 \\
 Transfusion & From Scratch 7B & - & - & - & - & - & 8.70 & - & 0.63 & - \\
 LMFusion & LLaVA-Next 8B & 1603.7 & 72.1 & 72.5 & 41.7 & - & \cellcolor{green!10}8.20 & - & - & - \\
 Janus & DeepSeek-LLM 1.5B & 1338.0 & 69.4 & 63.7 & 30.5 & 34.3 & 8.53 & 10.10 & 0.61 & - \\
 JanusFlow & DeepSeek-LLM 1.5B & 1333.1 & 74.9 & 70.5 & 29.3 & 30.9 & - & 9.51 & 0.63 & 80.09 \\
 Janus-Pro-1B & DeepSeek-LLM 1.5B & 1444.0 & 75.5 & 68.3 & 36.3 & 39.8 & - & 14.33$^\ddagger$ & 0.73 & 82.63 \\
 Janus-Pro-7B & DeepSeek-LLM 7B & 1567.1 & 79.2 & 72.1 & 41.0 & 50.0 & - & 13.48$^\ddagger$ & \cellcolor{green!10}0.80 & \cellcolor{green!10}84.19 \\
        \midrule
        \our-B& LLaVA-ov 0.5B & 1238.0 & 58.5 & 66.6 & 31.4 & 29.1 & 8.91 & 6.28 & 0.74$^\dagger$ & 80.04 \\
        \our-L& Qwen2.5-VL 3B & 1574.3 & 78.6 & 73.8 & 53.1 & 63.2 & 8.87 & 6.35 & 0.78$^\dagger$ & 81.10 \\
        \our-XL& Qwen2.5-VL 7B & \cellcolor{green!10}1685.2 & \cellcolor{green!10}83.5 & \cellcolor{green!10}76.9 & \cellcolor{green!10}58.6 & \cellcolor{green!10}66.6 & 8.69 & \cellcolor{green!10}6.02 & 0.80$^\dagger$ & 82.05 \\
        
    \end{tabular}
    \caption{Quantitative results on multimodal understanding and generation benchmarks. We report the COCO FID with Stable Diffusion v1.5~\citep{sd1p5}, and other metrics with Sana~\citep{sana}. $^\dagger$ denotes rewritten prompts. $^\ddagger$ denotes results tested by us under the same settings. }
    \label{tab:results}
\end{table*}

\subsection{Image Understanding and Generation}
\label{sec:img_understanding_generation}
\begin{figure*}[!t]
    \centering
    \includegraphics[width=\linewidth]{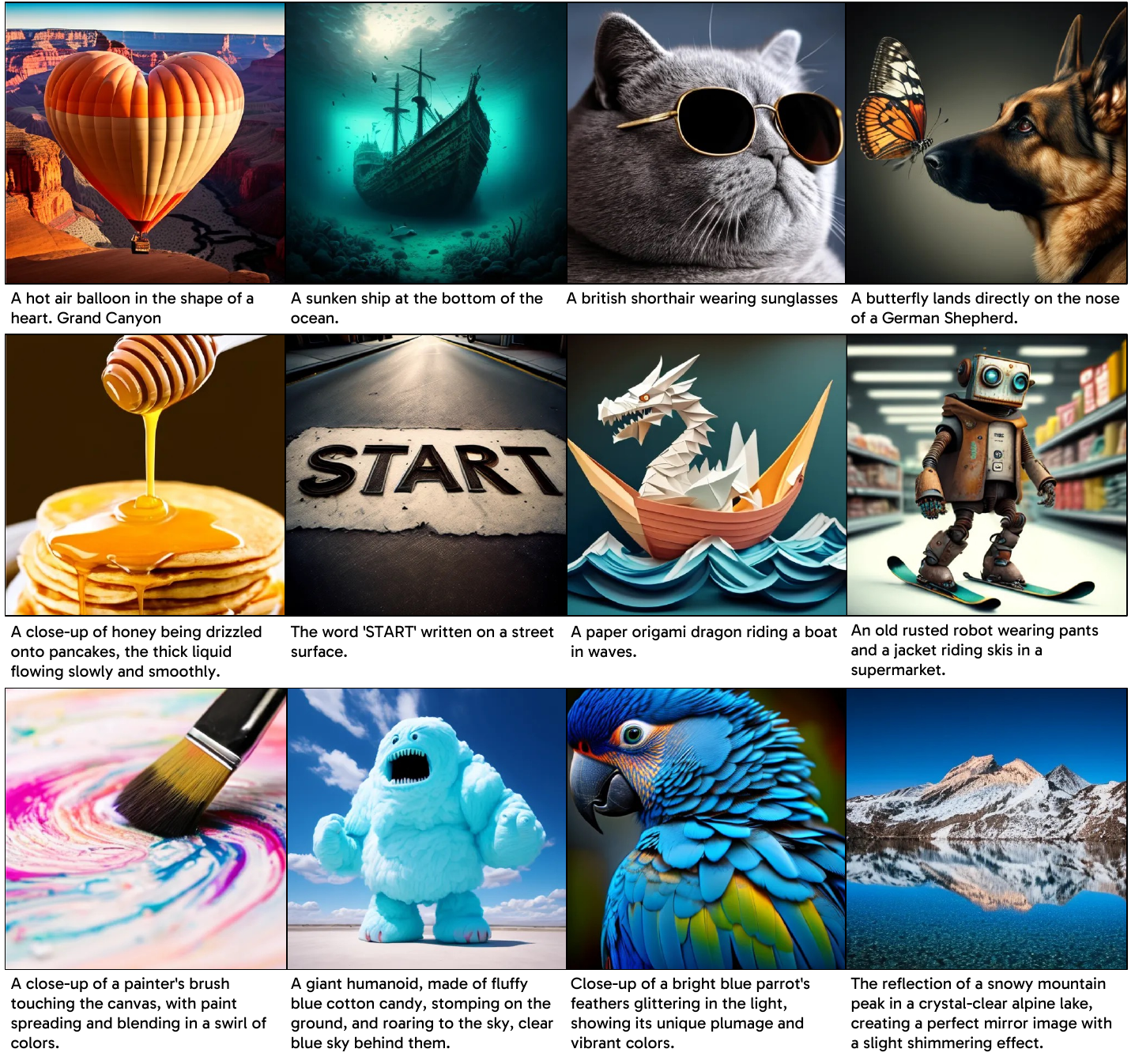}
    \caption{Qualitative results of \our. Prompts are from PartiPrompt~\citep{parti}, Sana~\citep{sana} and Movie Gen Bench~\citep{moviegen}.}
    \label{fig:qualitative}
\end{figure*}

As shown in Table~\ref{tab:results}, our model family demonstrates strong capabilities across both understanding and generation tasks. Benefiting from the flexible training approach that allows us to leverage arbitrary SOTA frozen MLLMs, all of our models in different sizes exhibit competitive performance on all understanding benchmarks~\citep{mme, mmbench, seedbench, mmmu, mmvet}. In terms of image generation, \our achieves SOTA visual quality on MJHQ-30K~\citep{playgroundv2p5}. Given the fact that \our works with frozen MLLMs, we can naturally connect with an arbitrary number of diffusion models. Since the base Sana-1.6B~\citep{sana} model is already fine-tuned on aesthetic data, we adopt Stable Diffusion v1.5~\citep{sd1p5} for COCO FID evaluation. Our results suggest that after adapting it to powerful MLLMs, we can achieve improved visual quality as indicated by the COCO FID score of 8.69. This also establishes a new SOTA COCO FID score among all Stable Diffusion v1.5-based unified models including MetaMorph~\citep{metamorph} (11.8) and Emu~\citep{emu} (11.66).

In terms of prompt alignment, \our also achieves competitive performance on GenEval~\citep{geneval} and DPG-Bench~\citep{dpg}, beating all diffusion model-based approaches including Transfusion~\citep{transfusion} and JanusFlow~\citep{janusflow}. We note that there is a performance gap between \our and Janus-Pro~\citep{januspro}, which auto-regressively generates image tokens. We suggest that this gap may be due to the different failure modes of diffusion models and auto-regressive models: diffusion models usually fail to correctly follow the prompt, while auto-regressive models may suffer from more visual artifacts, which are difficult to quantify by GenEval and DPG-Bench. We tested the MJHQ-30K FID score of Janus-Pro under the same setting as ours and found that, in terms of visual quality and artifact control, \our is significantly better than Janus-Pro (see Appendix~\ref{app:qualitative_comparison_with_sota_model} for visual comparison). Additionally, we find that \our achieves much better world knowledge reasoning capability than Janus-Pro, which we will elaborate on in Section~\ref{sec:reasoning_and_knowledge_augmented_generation}. We also found that when scaling up the size of frozen LLMs, the generation qulaity and prompt alignment also improves. \our provides a simple and principled way for leveraging the most advanced multimodal LLMs within a unified modeling framework. We also provide qualitative results in Figure~\ref{fig:qualitative} to illustrate the text-to-image generation capability of \our.

\begin{figure}[!t]
    \centering
    \begin{overpic}[width=0.9\linewidth]{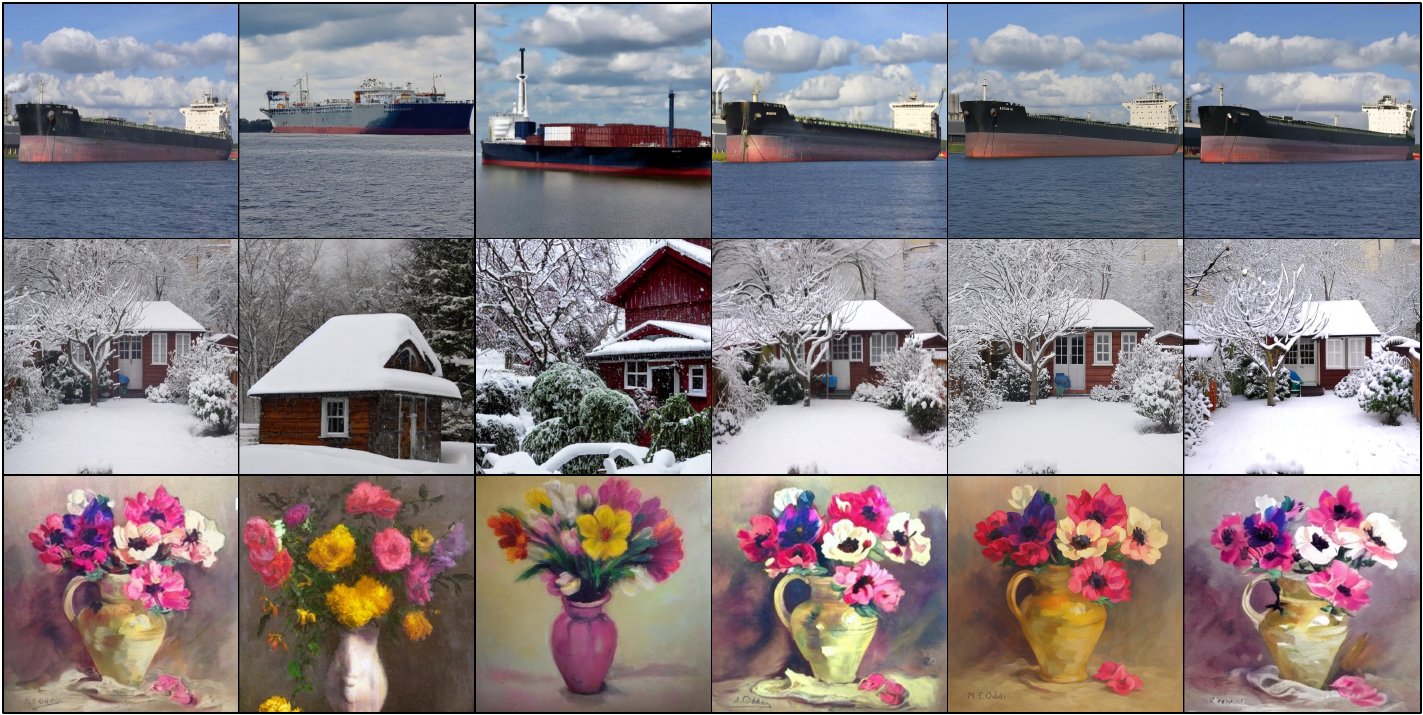}
        \put(1,-3) {\parbox{60pt}{\centering\scriptsize Real Image}}
        \put(18,-3) {\parbox{60pt}{\centering\scriptsize SEED\\\citep{seed}}}
        \put(35,-3.8) {\parbox{60pt}{\centering\scriptsize Emu\\\citep{emu}}}
        \put(51.5,-3.8) {\parbox{60pt}{\centering\scriptsize Emu2\\\citep{emu2}}}
        \put(68,-3) {\parbox{60pt}{\centering\scriptsize GPT-4o\\\citep{gpt4oimagegeneration}}}
        \put(85,-3) {\parbox{60pt}{\centering\scriptsize \our-B}}
    \end{overpic}
    \vspace{2em}
    \caption{Image reconstruction results. Results of SEED, Emu, and Emu2 are from \citet{emu2}.}
    \label{fig:image_reconstruction}
\end{figure}

\begin{figure}[!t]
    \centering
    \includegraphics[width=\linewidth]{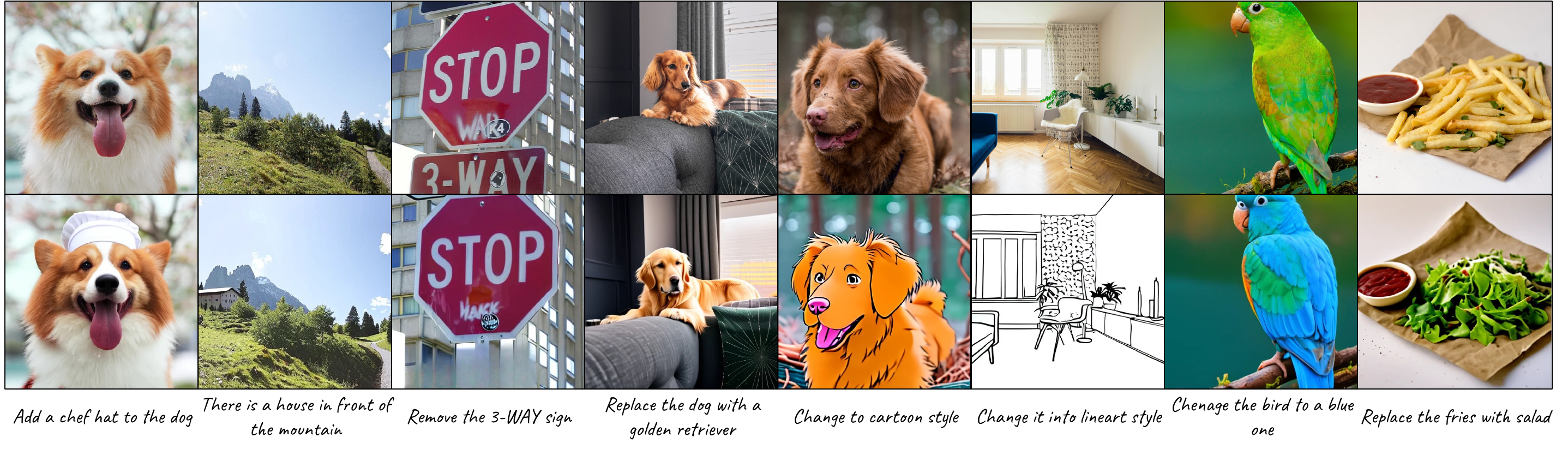}
    \caption{Image editing results. This capability can be easily transferred from image reconstruction after lightweight fine-tuning.}
    \label{fig:edit}
\end{figure}

\begin{figure}[!t]
    \centering
    \includegraphics[width=\linewidth]{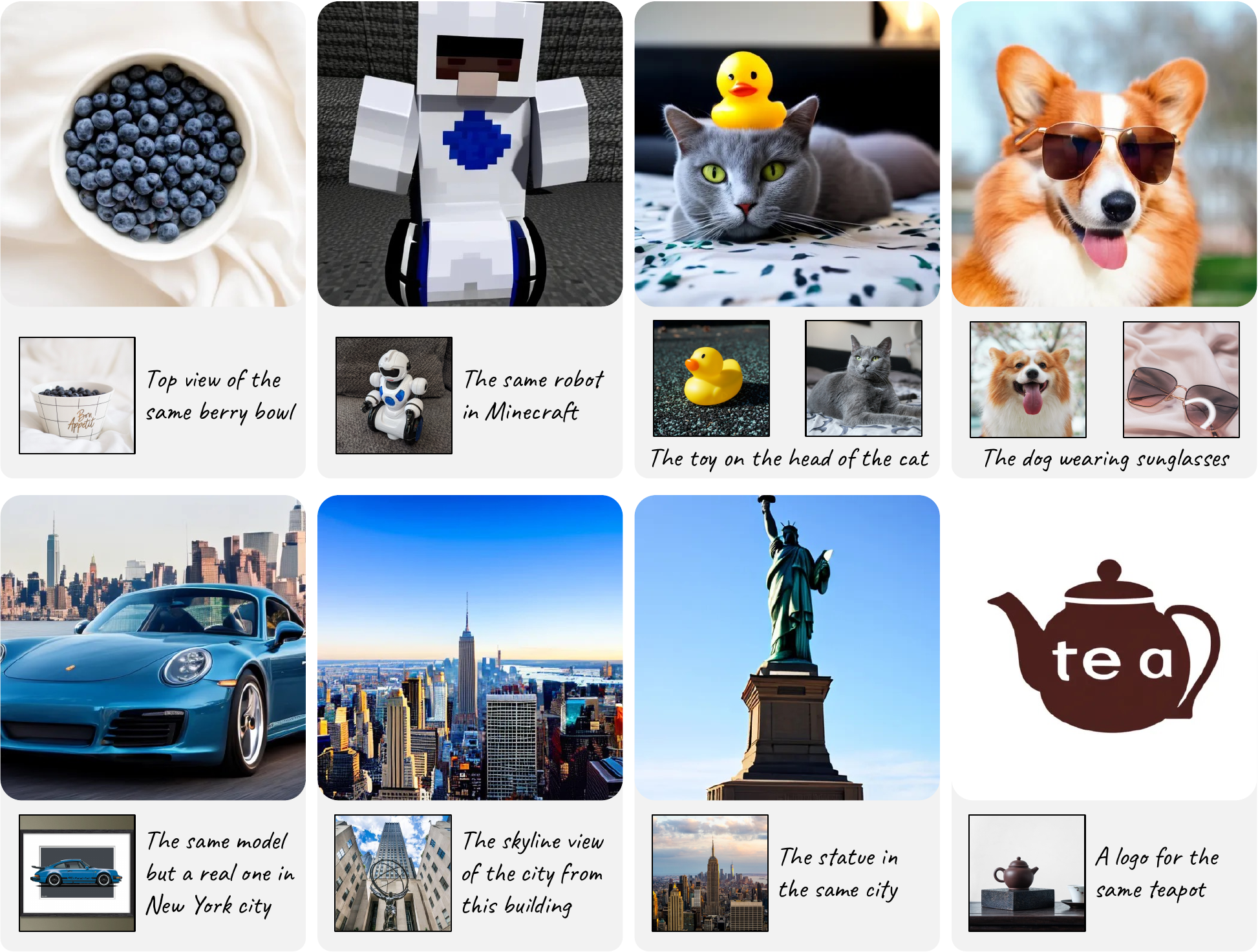}
    \caption{Qualitative results for instruction tuning. Instruction-tuned \our achieves strong subject-driven capability (first row) and can even reason through the multimodal input to generate images (second row).}
    \label{fig:subjectdriven}
\end{figure}

\begin{table}[!t]
    \small
    \centering
    \begin{tabular}{lccc}
 {\textbf{Methods}} & \textbf{DINO Score$\uparrow$} & \textbf{CLIP-I Score$\uparrow$} & \textbf{CLIP-T Score$\uparrow$} \\
    \midrule
 Real Images (Oracle) & 0.774 & 0.885 & -\\
    \midrule
    \multicolumn{4}{c}{\textit{fine-tuning}} \\
 Textual Inversion~\citep{textualinversion}& 0.569 & 0.780 & 0.255 \\
 DreamBooth~\citep{dreambooth} & 0.668 & 0.803 & \cellcolor{green!10}0.305 \\
 BLIP-Diffusion~\citep{blipdiffusion} & 0.670 & 0.805 & 0.302 \\
    \midrule
    \multicolumn{4}{c}{\textit{zero-shot \& test time tuning free}} \\
 Re-Imagen~\citep{reimagen} & 0.600 & 0.740 & 0.270 \\
 BLIP-Diffusion~\citep{blipdiffusion} & 0.594 & 0.779 & 0.300 \\
 Kosmos-G~\citep{kosmosg} & 0.694 & 0.847 & 0.287 \\
    \our-B-Instruct & \cellcolor{green!10}0.737 & \cellcolor{green!10}0.852 & 0.301 \\
    \end{tabular}
    \caption{Subject-driven generation results on DreamBench~\citep{dreambooth}.}
    \label{tab:dreambench}
\end{table}

\subsection{Image Reconstruction}
\label{sec:img_reconstruction}
We demonstrate that \our can be easily fine-tuned for image reconstruction tasks with a frozen MLLM (See Appendix~\ref{app:training_objectives} for more details). As shown in Figure~\ref{fig:image_reconstruction}, we compare our fine-tuned \our-B with existing diffusion autoencoders from various unified models, which reconstruct images from predicted visual features. Since these unified models are not explicitly fine-tuned for image reconstruction, their results are directly decoded from the vision encoder's output. Remarkably, even under this more constrained setup, our fine-tuned \our-B can still achieve competitive performance, matching the best existing open-source model Emu2~\citep{emu2}. When compared with GPT-4o~\citep{gpt4oimagegeneration}, our model also achieves comparable quality.

\subsection{Image Editing}
\label{sec:img_editing}
As shown in Figure~\ref{fig:edit}, we demonstrate that \our can transfer its image reconstruction capability to perform image editing. We keep the MLLM backbone frozen and fine-tune our pre-trained Base model for only 1,000 steps on publicly available image editing data. Qualitative results demonstrate that \our performs effectively in these image-editing scenarios.

\begin{figure}[!t]
    \centering
    \begin{overpic}[width=\linewidth]{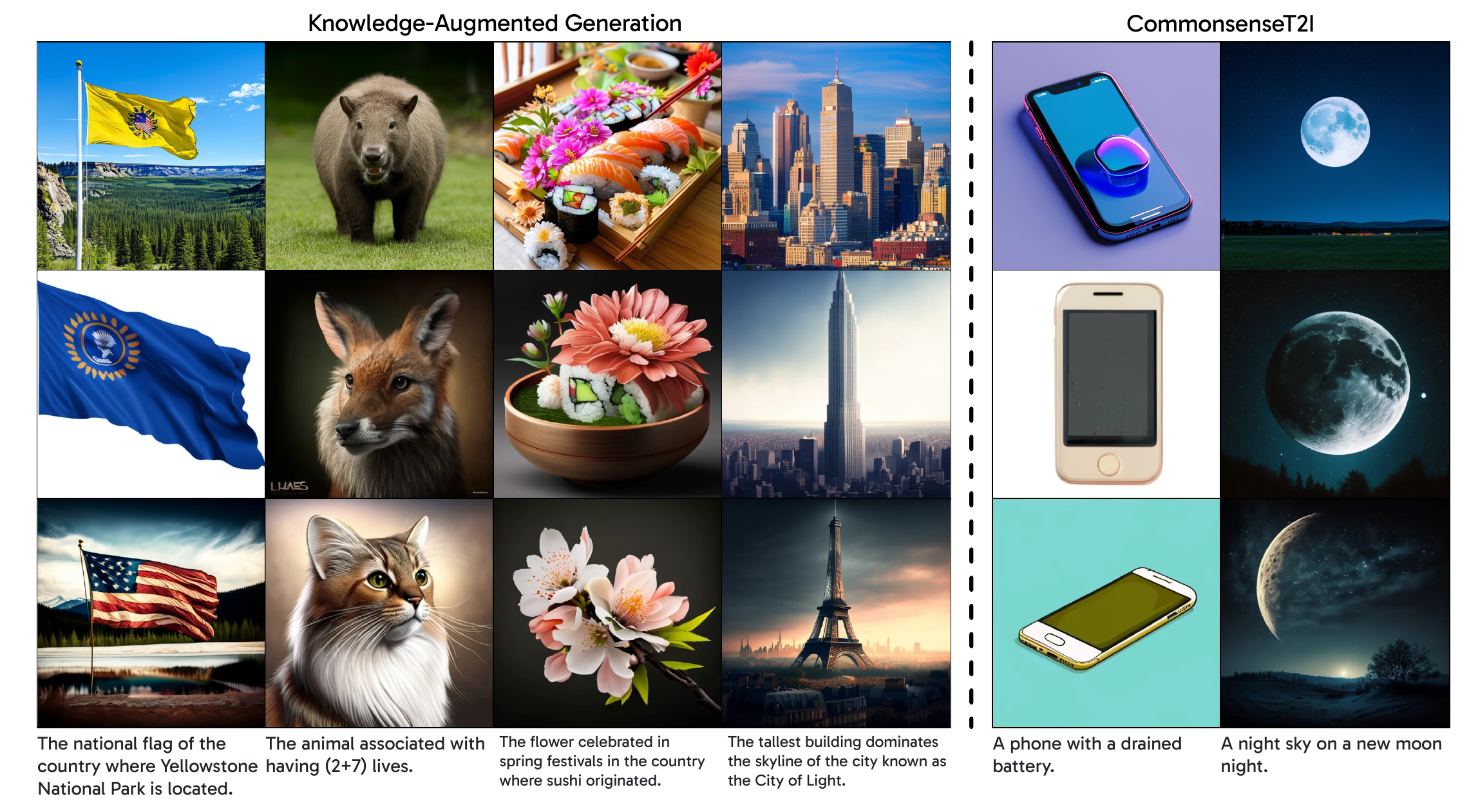}
        \put(-1,6) {\rotatebox{90}{\parbox{70pt}{\centering\scriptsize Ours-L w/\\\ours}}}
        \put(-1,22) {\rotatebox{90}{\parbox{70pt}{\centering\scriptsize Ours-L w/\\Last Layer Embed$^*$}}}
        \put(-1,38.5) {\rotatebox{90}{\parbox{62pt}{\centering\scriptsize Sana-1.6B\\\citep{sana}}}}
    \end{overpic}
    \caption{\our leverages frozen MLLMs for reasoning- and knowledge-augmented generation, overcoming the failure cases encountered in the base Sana model. $^*$ denotes that the LLM last layer embeddings of input tokens are used for image generation; the model is in L size (Qwen2.5-VL 3B). This approach can be better than the base Sana model in some cases but fails to activate in-context learning to perform knowledge-augmented generation. Some of the test cases are from MetaMorph~\citep{metamorph} and CommonsenseT2I~\citep{commonsenset2i}.}
    \label{fig:commonsense}
\end{figure}

\subsection{Instruction Tuning}
\label{sec:subject_driven}
We show that after being instruction-tuned on the proposed 2.4M dataset in Section~\ref{sec:model_training}, \our can achieve impressive zero-shot subject-driven generation performance, producing coherent results even with multiple highly customized subjects (the first row of Figure~\ref{fig:subjectdriven}). Using various supervision signals, the instruction-tuned \our-B model surprisingly unlocks novel capabilities like visual association and logo design that go beyond copy-pasting (the second row of Figure~\ref{fig:subjectdriven}). For example, in the first case, the model identifies the specific model of the input Porsche 911 car image, then correctly generates a novel front view for that model. In the second case, the model recognizes the input image of Rockefeller Center and imagines the view of New York City from the top of the Rockefeller Center.

We also follow DreamBooth~\citep{dreambooth} by adopting DINO, CLIP-I, and CLIP-T scores to quantitatively evaluate our model on the DreamBench~\citep{dreambooth} dataset. As shown in Table~\ref{tab:dreambench}, our \our-B-Instruct model achieves SOTA performance, outperforming existing models like Kosmos-G~\citep{kosmosg} that are explicitly trained on constructed substitution tasks for subject-driven generation.

\subsection{Reasoning- and Knowledge-Augmented Generation}
\label{sec:reasoning_and_knowledge_augmented_generation}

We show that the learnable queries can effectively leverage capabilities of the frozen LLM. This enables the model to better understand and follow complex prompts, including those requiring real-world knowledge and reasoning. As shown in Figure~\ref{fig:commonsense}, for the left knowledge-augmented generation cases, \our-L can leverage world knowledge from the frozen MLLM and reason through the input question to generate the correct answer. For the right commonsense knowledge cases from CommonsenseT2I~\citep{commonsenset2i}, the LLM provides better commonsense knowledge and enables \our to generate images that are consistent with the facts.

\begin{table}[!t]
    \centering
    \small
    \setlength{\tabcolsep}{8pt}
    \begin{tabular}{lccccccc}
    \textbf{Methods} & \textbf{Cultural}  & \textbf{Time}     & \textbf{Space}    & \textbf{Biology}    & \textbf{Physics} & \textbf{Chemistry} & \textbf{Overall} \\
    \midrule
    \rowcolor{green!25} GPT-4o$^{**}$~\citep{gpt4oimagegeneration} & \textbf{0.94} & \textbf{0.64} & \textbf{0.98} & \textbf{0.93} & \textbf{0.98} & \textbf{0.95} & \textbf{0.89} \\
    \midrule
    \multicolumn{8}{c}{\textit{Text-to-Image Models}} \\
 SD-v1-5~\citep{sd1p5} & 0.34 & 0.35& 0.32&0.28 &0.29 &0.21 & 0.32\\
 SD-XL~\citep{sdxl} &0.43  & 0.48 &0.47  &0.44  &0.45 &0.27 & 0.43 \\
 PixArt-Alpha~\citep{pixart} & 0.45  & 0.50& 0.48 & 0.49& 0.56 &0.34 & 0.47\\
 playground-v2.5~\citep{playgroundv2p5} & 0.49  &\cellcolor{green!10}0.58  & 0.55&0.43  & 0.48&0.33 & 0.49 \\
 SD-3.5-large~\citep{sd3} & 0.44 &0.50 &0.58  & 0.44&0.52 &0.31 & 0.46 \\
 FLUX.1-dev~\citep{flux} & 0.48  & \cellcolor{green!10}0.58 &\cellcolor{green!10}0.62  &0.42  &0.51 & 0.35 & 0.50 \\
    \midrule 
    \multicolumn{8}{c}{\textit{Unified Models}} \\
 show-o-512 \citep{showo} & 0.28 &0.40  &0.48 & 0.30& 0.46 & 0.30 & 0.35\\
 vila-u-7b-256 \citep{vilau} & 0.26 &0.33  & 0.37 &0.35  &0.39 &0.23 & 0.31\\
 Emu3~\citep{emu3} & 0.34 & 0.45 & 0.48 & 0.41  & 0.45 & 0.27 & 0.39 \\
 Janus-1.3B~\citep{janus} &0.16 &0.26 &0.35 & 0.28 &0.30 & 0.14& 0.23\\
 JanusFlow-1.3B~\citep{janusflow} &0.13 &0.26 &0.28 & 0.20& 0.19&0.11 & 0.18\\
 Janus-Pro-1B \citep{januspro} & 0.20& 0.28&0.45 & 0.24 & 0.32& 0.16& 0.26\\
 Janus-Pro-7B \citep{januspro} & 0.30& 0.37& 0.49 & 0.36&0.42 &0.26 & 0.35 \\
    \midrule
    \our-B & 0.44 & 0.49 & 0.58 & 0.41 & 0.49 & 0.34 & 0.46 \\
    \our-L & \cellcolor{green!10}0.56& 0.57 &\cellcolor{green!10}0.62 & 0.48 &  \cellcolor{green!10}0.63 &  \cellcolor{green!10}0.42 & \cellcolor{green!10}0.55 \\
    \our-XL & \cellcolor{green!10}0.56& 0.55 &\cellcolor{green!10}0.62 &  \cellcolor{green!10}0.49 &  \cellcolor{green!10}0.63 & 0.41 & \cellcolor{green!10}0.55 \\
    \end{tabular}
    \caption{Comparison of world knowledge reasoning on WISE~\citep{wise}. The test cases in WISE are similar to the knowledge-augmented generation ones in Figure~\ref{fig:commonsense}. \our achieves SOTA performance and significantly outperforms all other unified models. $^{**}$ Results are evaluated by \citet{gpt-imgeval} on a random subset of 200 out of 1000 samples.}
    \label{tab:wisescore}
\end{table}

\begin{table}[!t]
    \small
    \centering
    \begin{tabular}{lcc}
        \textbf{Methods} & \textbf{w/o Neg. Prompt} & \textbf{w/ Neg. Prompt} \\
        \midrule
 DALL-E 3~\citep{dalle} w/ rewrite & \cellcolor{green!10}40.17 & N/A \\
 SD-XL~\citep{sdxl} & 26.00 & 44.83 \\
 SD-3-medium~\citep{sd3} & 26.17 & 47.17 \\
 FLUX.1-dev~\citep{flux} & 24.50 & 22.50 \\
 Sana-1.6B~\citep{sana} & 25.17 & 43.33 \\
        % \our-B Last Layer Embed$^*$ & 26.33 & 48.67 \\
        \our-B & 27.33 & 51.50 \\
        \our-L & 28.83 & \cellcolor{green!10}57.67 \\
        
    \end{tabular}
    \caption{Comparison of visual commonsense reasoning capability on CommonsenseT2I~\citep{commonsenset2i}.}
    \label{tab:commonsenset2i}
\end{table}

To quantitatively evaluate \our's world knowledge reasoning capability, we employ the WISE~\citep{wise} benchmark, which contains similar test cases to the knowledge-augmented generation examples shown in Figure~\ref{fig:commonsense}. As demonstrated in Table~\ref{tab:wisescore}, \our achieves SOTA performance, significantly outperforming all other unified models. Notably, before our work, existing unified models struggled to effectively leverage powerful MLLMs for reasoning and knowledge-augmented generation, resulting in inferior performance compared to text-to-image models. \our stands as the first unified model to successfully transfer the advanced capabilities of frozen MLLMs to image generation and exceed the performance of SOTA text-to-image models.

We also quantitatively evaluate \our's commonsense reasoning capability on the CommonsenseT2I benchmark~\citep{commonsenset2i} in Table~\ref{tab:commonsenset2i}. For simplicity, we use CLIP~\citep{clip} as the evaluator following their original implementation. Results show that \our significantly improves the performance of the base Sana model, achieving SOTA performance.

\subsection{Discussion}
\label{sec:discussion}

\paragraph{Comparison over different LLM backbones.}
As shown in Table~\ref{tab:llm_comparsion}, to test the impact of employing different LLM backbones for \our, we carefully select a family of backbone models: pre-trained LLM (Qwen2.5-3B), instruction-tuned LLM (Qwen2.5-3B-Instruct), and instruction-tuned MLLM (Qwen2.5-VL-3B-Instruct). Both instruction-tuned models are initialized with the first pre-trained model checkpoint. Experimental results show that instruction tuning can achieve better (multimodal) understanding capabilities. However, the improvements are orthogonal to image generation performance when employed to provide multimodal generation conditions.

\paragraph{Comparison with using last layer embeddings.}
As shown in Table~\ref{tab:learnable_queries}, our learnable queries approach achieves comparable image generation quality and prompt alignment to using the LLM's last layer embeddings of input tokens. However, the last layer embedding method essentially treats the decoder-only LLM as a text encoder, which inherently limits its in-context learning capabilities. While this approach does improve upon the base Sana model in some cases as demonstrated in Figure~\ref{fig:commonsense}, it struggles with the knowledge-augmented generation cases shown in the same figure. These cases require the LLM to first process and answer input questions before generating corresponding images, demanding in-context learning beyond what text encoders typically provide. This performance gap is quantitatively confirmed in Table~\ref{tab:encoder_style_comparsion}, where \our significantly outperforms the last layer embedding approach on both WiScore and CommonsenseT2I benchmarks. Integrated natively with the LLM, \our naturally leverages its in-context learning capabilities, enabling the model to reason through questions and generate appropriate images.

\begin{table}[!t]
    \small
    \centering
    \begin{tabular}{lcccc}
        \textbf{LLM Backbones} & \textbf{MJHQ-30K FID $\downarrow$} & \textbf{GenEval $\uparrow$} & \textbf{DPG-Bench $\uparrow$} & \textbf{CommonsenseT2I $\uparrow$} \\
        \midrule
 Qwen2.5-3B & 6.20 & 0.79 & 81.34 & 56.00 \\
 Qwen2.5-3B-Instruct & 6.36 & 0.79 & 81.12 & 54.33 \\
 Qwen2.5-VL-3B-Instruct & 6.35 & 0.78 & 81.10 & 57.67 \\
        
    \end{tabular}
    \caption{Comparison across different LLM backbones. Image generation capability is mostly orthogonal to multimodal understanding capability.}
    \label{tab:llm_comparsion}
\end{table}

\begin{table}[!t]
    \small
    \centering
    \begin{tabular}{lccccc}
        \textbf{Methods} & \textbf{MJHQ-30K FID $\downarrow$} & \textbf{GenEval $\uparrow$} & \textbf{DPG-Bench $\uparrow$} & \textbf{WiScore $\uparrow$} & \textbf{CommonsenseT2I $\uparrow$} \\
        \midrule
 Ours-L w/ Last Layer Embed$^*$ & 6.41 & \cellcolor{green!10}0.78 & \cellcolor{green!10}81.23 & 0.48 & 52.83 \\
 Ours-L w/ \ours & \cellcolor{green!10}6.35 & \cellcolor{green!10}0.78 & 81.10 & \cellcolor{green!10}0.55 & \cellcolor{green!10}57.67 \\
    \end{tabular}
    \caption{Comparison between \our and LLM last layer embedding. $^*$ denotes that the LLM last layer embeddings of input tokens are used for image generation. We observe comparable performance between \our and LLM last layer embedding on visual quality and prompt alignment. However, \our can activate in-context learning to perform knowledge-augmented generation, yielding much better performance on commonsense reasoning and world knowledge reasoning.}
    \label{tab:encoder_style_comparsion}
\end{table}

\section{Conclusion}
We presented \ours, a simple interface connecting MLLMs (for understanding) and diffusion decoders (for generation), effective even when the MLLM is frozen. This approach yields state-of-the-art understanding and generation performance with straightforward implementation. By enabling transfer between modalities, \ours successfully channels MLLM knowledge and reasoning into multimodal generation. While effective, we hypothesize that bridging the remaining gap to leading proprietary systems may primarily involve further data scaling. We hope \ours provides a powerful, accessible baseline for future unified multimodal model development.

\clearpage
\newpage
\bibliographystyle{assets/plainnat}
\bibliography{paper}

\clearpage
\newpage
\beginappendix
\section{Data Curation Details}
\label{app:details_in_data_curation}
For the data curation part, we use \texttt{Qwen/Qwen2-VL-7B-Instruct}\footnote{\url{https://huggingface.co/Qwen/Qwen2-VL-7B-Instruct}} as our MLLM, The system prompt we are using is:
\canvas{
Based on the provided of one or multiple source images, one target image, and their captions, create an interesting text prompt that can be used with the source images to generate the target image.

This prompt should include:
\begin{itemize}
\item one general and unspecific similarity shared with the source images (same jersey top, similar axe, similar building, etc).
\item all differences that only the target image has.
\end{itemize}

This prompt should NOT include:
\begin{itemize}
\item any specific details that would allow generating the target image independently without referencing the source images.
\end{itemize}

Remember the prompt should be concise and short. The generation has to be done by combining the source images and text prompts.
}

\section{Qualitative Comparison with SOTA Open-Source Model on Text-to-Image Generation}
\label{app:qualitative_comparison_with_sota_model}
\begin{figure}[!t]
    \centering
    \begin{overpic}[width=\linewidth]{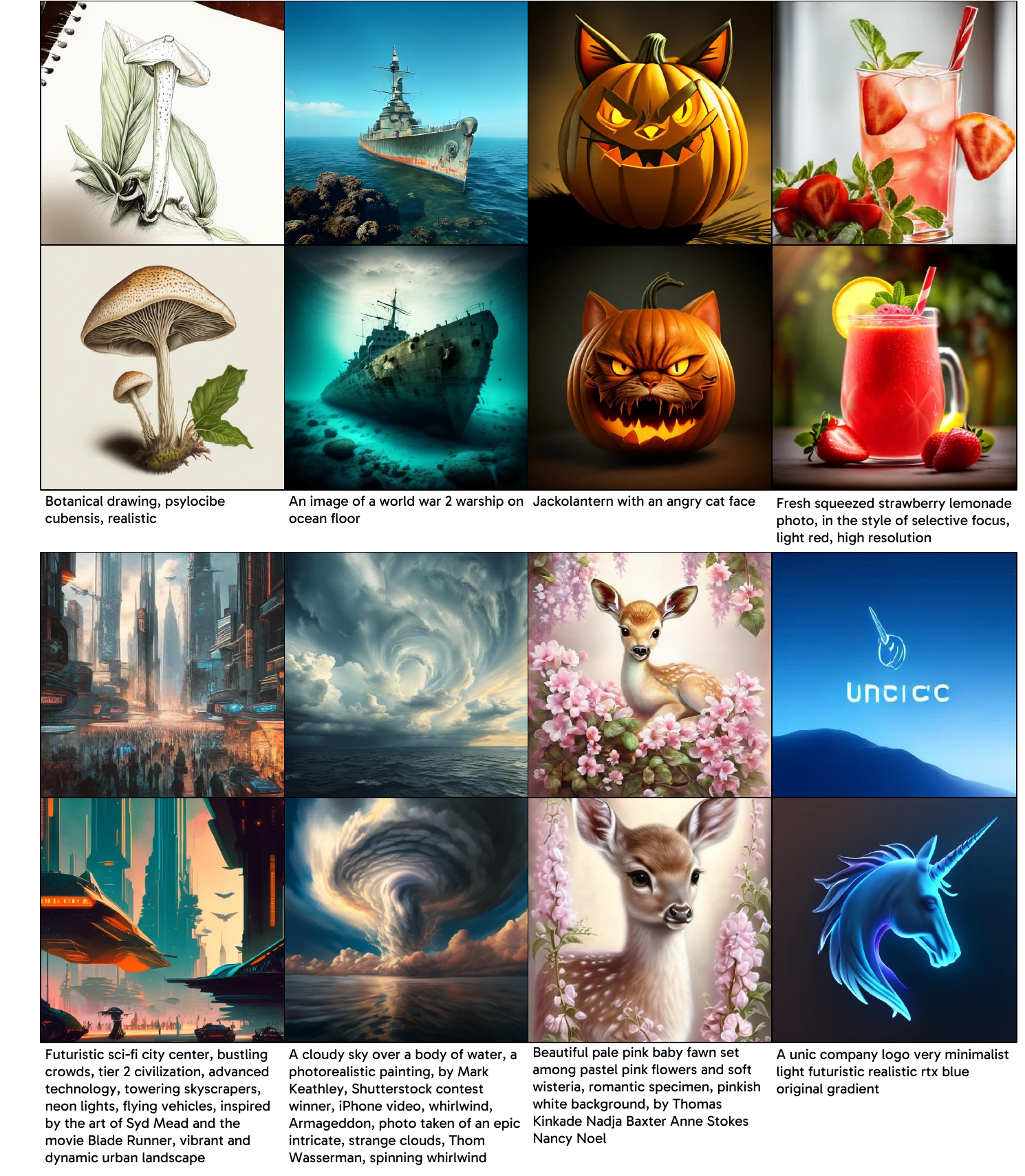}
        \put(0,16) {\rotatebox{90}{\small \our-XL}}
        \put(-1,33) {\rotatebox{90}{\parbox{100pt}{\centering \small Janus-Pro-7B\\\citep{januspro}}}}
        \put(0,63) {\rotatebox{90}{\small \our-XL}}
        \put(-1,80) {\rotatebox{90}{\parbox{100pt}{\centering \small Janus-Pro-7B\\\citep{januspro}}}}
    \end{overpic}
    \caption{Qualitative comparison with Janus-Pro-7B~\citep{januspro} on MJHQ-30K~\citep{playgroundv2p5}.}
    \label{fig:compare_janus}
\end{figure}

We provide a qualitative comparison with Janus-Pro-7B~\citep{januspro} on MJHQ-30K~\citep{playgroundv2p5} in Figure~\ref{fig:compare_janus}. We can see that \our-XL follows the prompt better and generates more visually appealing images than Janus-Pro-7B.

\section{Training Objectives}
\label{app:training_objectives}

\begin{table}[!ht]
    \small
    \centering
    \begin{tabular}{lcccc}
        \textbf{Objective} & \textbf{Rel. Wall Time} & \textbf{MJHQ-30K FID $\downarrow$} & \textbf{GenEval $\uparrow$} & \textbf{DPG-Bench $\uparrow$} \\
        \midrule
 Text-to-Image & 1.0x & \cellcolor{green!10}7.43 & \cellcolor{green!10}0.56 & 75.35 \\
 Image Reconstruction & 2.79x & 27.42 & 0.32 & 68.36 \\
 Mix & 2.61x & 8.27 & 0.54 & \cellcolor{green!10}76.53 \\
    \end{tabular}
    \caption{Study on training objectives. Image reconstruction objective can be mixed with text-to-image objective to enable image reconstruction capabilities without harming visual quality and prompt alignment.}
    \label{tab:objective}
\end{table}

We are using an MLLM for multimodal perception, besides the standard text-to-image objective, we can also use an image reconstruction objective to achieve alignment. In Table~\ref{tab:objective}, we show that training with the text-to-image objective achieves much better performance than the image reconstruction objective. We demonstrate that a mix of both objectives can enable image reconstruction capabilities without being generally harmful to the T2I performance.

\end{document}